\documentclass[letterpaper]{article} 
\usepackage{aaai24}  
\usepackage{times}  
\usepackage{helvet}  
\usepackage{courier}  
\usepackage[hyphens]{url}  
\usepackage{graphicx} 
\usepackage{natbib}  
\usepackage{caption} 
\usepackage{algorithm}
\usepackage{algorithmic}

\usepackage{newfloat}
\usepackage{listings}
\DeclareCaptionStyle{ruled}{labelfont=normalfont,labelsep=colon,strut=off} 
\floatstyle{ruled}
\newfloat{listing}{tb}{lst}{}
\floatname{listing}{Listing}
\usepackage{amsmath}
\usepackage{booktabs}      
\usepackage{subcaption} 
\usepackage{amsfonts}
\usepackage{mathtools} 

\title{Pantypes: Diverse Representatives for Self-Explainable Models}

\author{
    Rune Kjærsgaard \textsuperscript{\rm 1}, Ahcène Boubekki \textsuperscript{\rm 2}, Line Clemmensen \textsuperscript{\rm 1}\\
}

\affiliations{
    \textsuperscript{\rm 1} Department of Applied Mathematics and Computer Science, Technical University of Denmark, Denmark\\
    \textsuperscript{\rm 2} Machine Learning and Uncertainty, Physikalisch-Technische Bundesanstalt, Germany

    rdokj@dtu.dk
}

\begin{document}

\maketitle

\begin{abstract}
Prototypical self-explainable classifiers have emerged to meet the growing demand for interpretable AI systems. These classifiers are designed to incorporate high transparency in their decisions by basing inference on similarity with learned prototypical objects. While these models are designed with diversity in mind, the learned prototypes often do not sufficiently represent all aspects of the input distribution, particularly those in low density regions. 
Such lack of sufficient data representation, known as representation bias, has been associated with various detrimental properties related to machine learning diversity and fairness. In light of this, we introduce \emph{pantypes}, a new family of prototypical objects designed to capture the full diversity of the input distribution through a sparse set of objects. We show that pantypes can empower prototypical self-explainable models by occupying divergent regions of the latent space and thus fostering high diversity, interpretability and fairness.
\end{abstract}

\section{Introduction}
Machine learning (ML) systems are increasingly affecting individuals across various societal domains. This has put into question the black-box nature of these systems, and fostered the field of explainable AI (XAI), wherein model inference is corroborated with justifications and explanations in an effort to increase transparency and trustworthiness. In this line of research two approaches have arisen; that of ad-hoc black-box model explanations \citep{selvaraju2017grad,yosinski2015understanding}, and that of self-explainable models (SEMs) \citep{chen2019looks,alvarez2018towards}. A popular approach for SEMs substitutes traditional black-box networks with glass-box counterparts, where class representative prototypes are generated and used in the decision process \citep{chen2019looks} leading to increased trustworthiness and interpretability.

The various initiatives emerging in the literature share the same overarching goals, but there is still a lack of consensus on the exact properties a SEMs should display \citep{gautam2023looks}. We adopt three prerequisites of a SEM outlined in \cite{gautam2022protovae}, namely \textit{transparency}, \textit{trustworthiness} and  \textit{diversity}.

\textit{Transparency} may be defined by two properties; (i) the learned concepts are used in the decision making process without the use of a black-box model and 
(ii) the learned concepts can be visualized in the input space. 

\textit{Trustworthiness} may be defined by three properties; (i) the predictive performance of the model matches its closest black-box counterpart, (ii) explanations are robust and (iii) the explanations directly represent the contribution of the input features to the model predictions.

\textit{Diversity} may be defined by one property; (i) the concepts learned by the SEM are represented by non-overlapping information in the latent space.

While significant work has been put forth in the literature to cement the transparency and trustworthiness axis of SEMs, only limited effort using qualitative measures exists for the diversity axis. Similarly, the relation between the diversity axis and appropriate inference remains largely unexplored. Diversity is typically ensured by introducing model regularization towards learning non-overlapping concepts \cite{vilone2020explainableDiversity}. However, this condition may not be strong enough, as non-overlapping concepts can still be learned in a small region of the input space, causing a lack of representativity for the full data distribution, known as representation bias \citep{shahbazi2022survey}. Representation bias can cause smaller sub-populations to remain hidden in low-density regions and ultimately cause biased inference \citep{jin2020mithracoverage}. To provide sufficient coverage and to mitigate the impact of data bias during model inference, it is critical to capture the full diversity of the data, and to have this diversity be represented in the prototypes learned by the SEM. To this end, we introduce pantypes, a new family of prototypical objects designed to empower SEMs by sufficiently covering the dataspace. Pantype generation is promoted using a novel volumetric loss inspired by a probability distribution known as a Determinantal Point Process (DPP) \citep{kulesza2012determinantal}. This loss induces higher prototype diversity, enables more fine-grained diversity control, and at the same time allows prototype pruning wherein the number of prototypes is determined dynamically dependent on the diversity expressed within each class. Prototype pruning enables the capacity to learn additional prototypes for complex classes and to grasp simple classes through a sparser set of objects, improving the interpretability of the class representatives. 

Our contributions can be summarized as follows: 
\begin{itemize}
    \item Introduction of a volumetric loss, which promotes the generation of pantypes, a highly diverse set of prototypes.
    \item Quantitative measures for prototype representativity and diversity in SEMs.
    \item Dynamic class-specific prototype selection.
\end{itemize}

\section{PanVAE}
The modeling task at hand involves a classification setting on visual image data, where the SEM learns to classify $K > 0$ classes from a training set $\mathbf{X} = \{(\boldsymbol{x_i},\boldsymbol{y_i})\}^N_{i=1}$ ,where $\boldsymbol{x_i} \in \mathbb{R}^P$ is the $i^\textup{th}$ image and $\boldsymbol{y_i} \in \{0,1\}^K$ is a one-hot label vector.
We implement the pantypes\footnote{Our code and training details are publicly available on GitHub at https://github.com/RuneDK93/pantypes} on the foundation of a well-tested variational autoencoder based SEM, known as ProtoVAE \citep{gautam2022protovae}. This model uses an encoder function $f : \mathbb{R}^p \rightarrow \mathbb{R}^d \times \mathbb{R}^d$, to transform the input images into a posterior distribution $(\boldsymbol{\mu_i},\boldsymbol{\sigma_i})$. A latent representation $\boldsymbol{z_i}$ of the $i^\textup{th}$ image is then sampled from the distribution $\mathcal{N}(\boldsymbol{\mu_i},\boldsymbol{\sigma_i})$ and passed as input to a decoder function $g : \mathbb{R}^d \rightarrow \mathbb{R}^p$ to generate the reconstructed image $g(\boldsymbol{z_i}) = \hat{\boldsymbol{x_i}}$. To enable transparent predictions, the model does not directly use the feature vector $\boldsymbol{z_i}$ during inference, but rather compares this vector to $M$ prototypes per class $\boldsymbol{\Phi} = \{\phi_{kj}\}^{k=1..K}_{j=1,,M}$ via a similarity function : $R^d \rightarrow R^M$. The resulting similarity vector $\boldsymbol{s}_i \in \mathbb{R}^{K \times M}$ is then used in a glass-box linear classifier $h : \mathbb{R^M} \rightarrow [0,1]^K$ to generate the class prediction $h(\boldsymbol{s_i}) = \hat{\boldsymbol{y}_i}$. The similarity function \citep{chen2019looksSIMfunc} is given by:
\begin{equation}
\label{EQ:sim_function}
    \boldsymbol{s}_i(k,j) = \mathrm{sim}(\boldsymbol{z}_i,\boldsymbol{\phi_{kj}}) = \mathrm{log} \left( \frac{||\boldsymbol{z}_i - \boldsymbol{\phi_{kj}}||^2 + 1 }{||\boldsymbol{z}_i-\boldsymbol{\phi_{kj}}||^2 + \epsilon} \right),
\end{equation}
where $0 < \epsilon < 1$. This construction allows the similarity vector to not only capture the distances to the prototypes, but to also reflect the influence of each prototype on the final prediction. 

\subsection{Loss Terms}
To further enforce the properties of a SEM, we adopt the same prediction and VAE loss term structure as ProtoVAE:
\begin{equation}
    \mathcal{L}_\mathrm{ProtoVAE} =  \mathcal{L}_\mathrm{pred} + \mathcal{L}_\mathrm{VAE} + \mathcal{L}_\mathrm{orth},
\end{equation}
where 
\begin{equation}
    \mathcal{L}_\mathrm{pred} = \frac{1}{N}\sum^N_{i=1} \mathrm{\textbf{CE}}(h(\boldsymbol{s}_i);\boldsymbol{y}_i)
\end{equation}    
is a cross-entropy (CE) prediction loss term ensuring inter-class diversity in the prototypes and 
\begin{equation}
\begin{multlined}
    \mathcal{L}_\mathrm{VAE} = \frac{1}{N}\sum^N_{i=1}  || \boldsymbol{x}_i - \hat{\boldsymbol{x}_i}||^2 + \sum^K_{k=1 }\sum^M_{j=1} \\ \boldsymbol{y}_i(k) \frac{\boldsymbol{s}_i(k,j)}{\sum^M_{l=1}\boldsymbol{s}_i(k,j)} D_\mathrm{KL} ( \mathcal{N}(\boldsymbol{\mu_i},\boldsymbol{\sigma_i}) || \mathcal{N}(\boldsymbol{\phi_{kj}},\boldsymbol{\boldsymbol{I}}_d)) 
\end{multlined}
\end{equation}    
is the loss for a mixture of VAEs using the same network each with a Gaussian prior distribution centered on one of the prototypes \citep{gautam2022protovae}. Here $\boldsymbol{\boldsymbol{I}}_d$ is a $d \times d$ identity matrix. 
Finally, an orthonormality loss term is used: 
\begin{equation}
    \mathcal{L}_\mathrm{orth} = \sum^K_{k=1}  || \boldsymbol{\bar{\Phi}}_k^T \boldsymbol{\bar{\Phi}}_k - \boldsymbol{I}_M ||^2_F,
\end{equation}    
where $\boldsymbol{\bar{\Phi}}_k$ is the mean subtracted prototype vector for all prototypes of class $k$ and $\boldsymbol{\boldsymbol{I}}_M$ is an $M \times M$ identity matrix. 

The orthonormality loss is included to foster intra-class prototype diversity and to uphold the diversity property of a SEM by inducing the learning of non-overlapping concepts in the latent space and thus avoiding prototype collapse \citep{wang2021interpretableProtoCollapse1,jing2021understandingProtoCollapse2}. While this loss causes the prototypes to be orthogonal, it does not explicitly prevent the prototypes from occupying and representing a small region (volume) of the full data-space. Moreover, prototype orthonormality is typically achieved early during training, and further scaling of the orthonormality loss does not significantly alter the diversity of the prototypes (see results section). 

Poor or skewed data representation, known as representation bias, has been associated with various detrimental properties related to ML fairness, where underrepresented minority groups are negatively affected during inference \cite{phillips2011otherBiasFace}. To mitigate these issues it is essential to achieve sufficient coverage of the full diversity represented in the data \cite{suresh2019frameworkcoverage}. We draw on this idea to empower the ProtoVAE model by exchanging its class-wise orthonormality diversity loss with a volumetric diversity loss, which causes the model to learn prototypical objects with various improved qualities, including an improved coverage of the embedding space. We call these learned objects \textit{pantypes}. The loss term structure of our model is:
\begin{equation}
    \mathcal{L}_\mathrm{PanVAE} =  \mathcal{L}_\mathrm{pred} + \mathcal{L}_\mathrm{VAE} + \mathcal{L}_\mathrm{vol},
\end{equation}
where $\mathcal{L}_\mathrm{vol}$ is the volumetric prototype loss, which not only prevents prevents prototype collapse, but causes higher prototype diversity, enables more fine-grained diversity control, and at the same time allows prototype pruning wherein the number of prototypes is determined dynamically dependent on the diversity expressed within each class.

\subsection{Pantypes}
Pantypes are prototypical objects learned in an end-to-end manner during model training. They are inspired by a probability distribution known as a Determinantal Point Process (DPP) \citep{kulesza2012determinantal}, which can be used to sample from a population while ensuring high diversity. DPPs have recently garnered attention in the ML community, and have been used to draw diverse sets in a range of ML applications including data from videos, images, documents, sensors and recommendation systems  \citep{dppvideo,kulesza2012determinantal,dppdocument,dpprecommender,dppsensor}. DPPs describe a distribution over subsets, such that the sampling probability of a subset is proportional to the determinant of an associated sub-matrix (a minor) of a positive semi-definite kernel matrix. The kernel matrix expresses similarity between feature vectors of observations through a kernel function $\boldsymbol{G}_{ij} = g(\boldsymbol{v}_i,\boldsymbol{v}_j)$. This global measure of similarity is then used to sample such that similar items are unlikely to co-occur. The kernel can be constructed in various ways including the radial basis function (RBF) kernel $\boldsymbol{G}_{ij} = e^{{-\gamma ||\boldsymbol{v}_i-\boldsymbol{v}_j||^2}}$ or the linear kernel, leading to a similarity function of inner products known as the Gram matrix  $\boldsymbol{G}_{ij} =  \langle \boldsymbol{v}_i,\boldsymbol{v}_j  \rangle$. When using the Gram matrix, a DPP is equivalent to sampling with probability proportional to the volume of the paralellotope formed by the feature vectors of the sampled items. We utilize the linear kernel to construct a volumetric loss on the prototypes in the following way: 
\begin{equation}
    \mathcal{L}_\mathrm{vol} =  \frac{1}{K}\sum^K_{k=1} \frac{1}{| \boldsymbol{G}_k |^{\frac{1}{2}}},
\end{equation}
where $\boldsymbol{G}_k \in \mathbb{R}^{M \times M}$ is the Gram matrix given by $\boldsymbol{G}_k = \boldsymbol{\Phi}_k^T \boldsymbol{\Phi}_k $ with $\Phi_{kj}$ as column vectors in $\boldsymbol{\Phi}_k$ and $|\boldsymbol{G}_k|$ is the Gramian (Gram determinant). $|\boldsymbol{G}_k|^{\frac{1}{2}}$ measures the $M$-dimensional volume of the parallelotope formed by the $M$ columns of $\boldsymbol{\Phi}_{k}$ embedded in $d$-dimensional space. In other words, it expresses the diversity of the $M$ prototypes of class $k$ through the volume spanned by their feature vectors. This loss not only prevents prototype collapse by causing the feature vectors to diverge, but also directly encourages the pantypes to occupy different sectors of the data domain to express a large volume.

\subsubsection{Prototype elimination} 
Increasing the scaling on the volume loss punishes pantypes that express a low volume and thus directly alters the diversity of the learned objects. With sufficient scaling, the volumetric loss forces pantypes out-of-distribution (OOD) if they are not necessary to represent the observed diversity of a class. This allows natural pruning, wherein the number of pantypes can be dynamically tuned by elimination of OOD pantypes. This is similar to the discipline of hyperspectral endmember unmixing, where a number of endmembers (prototypes) are disentangled from a hyperspectral image and linear combinations of the endmembers are used to reconstruct the input images. Following training, the learned endmembers can be associated with purity scores \citep{berman2004ice}, which express the quality of their explanations. These scores describe the maximal responsibility proportion of endmembers for reconstructing the original images. In other words, a high purity score indicates that an endmember shares a high similarity with individual input images, while a low purity score indicates that an endmember is capturing noise and should be pruned. Such purity scores can be constructed from the similarity scores used in the linear classifier in our SEM. Thus, as proposed by \citep{berman2004ice}, we can initiate the model with a sufficiently large number of pantypes, and use the similarity scores to prune individual OOD pantypes. We propose a heuristic for pruning, where a pantype can be pruned if it does not have the maximal similarity score for any of the training images (i.e. it does not individually represents any of the training images more than the other pantypes). 

\begin{figure*}[ht!]
    \centering
    \begin{subfigure}[t]{1\columnwidth}
        \includegraphics[width=1\columnwidth]{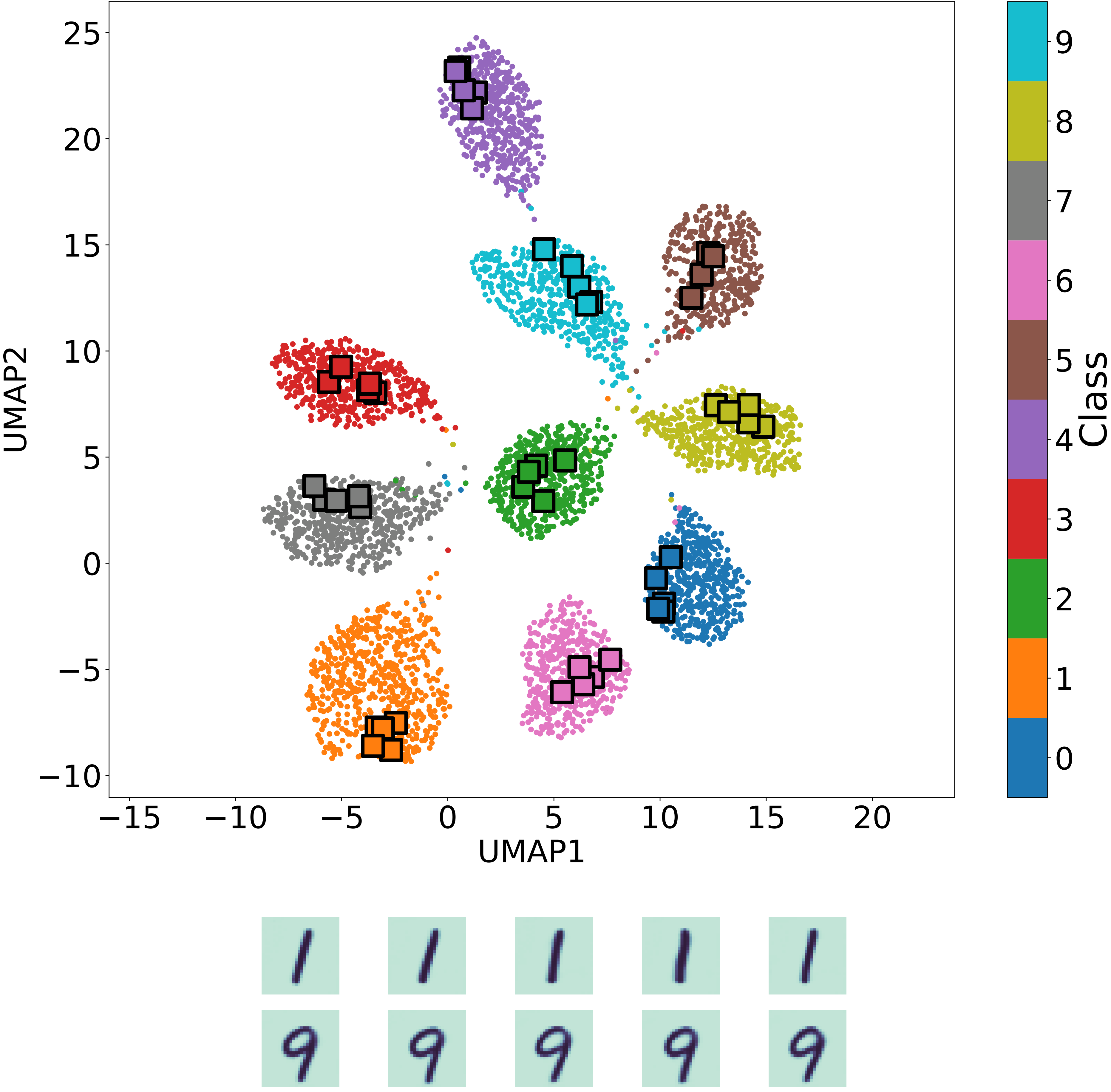}             
        \caption{ProtoVAE.}
        \label{fig:UMAP_MNIST30_ProtoVAE}    
    \end{subfigure}
    ~ 
    \begin{subfigure}[t]{1\columnwidth}    
        \includegraphics[width=1\columnwidth]{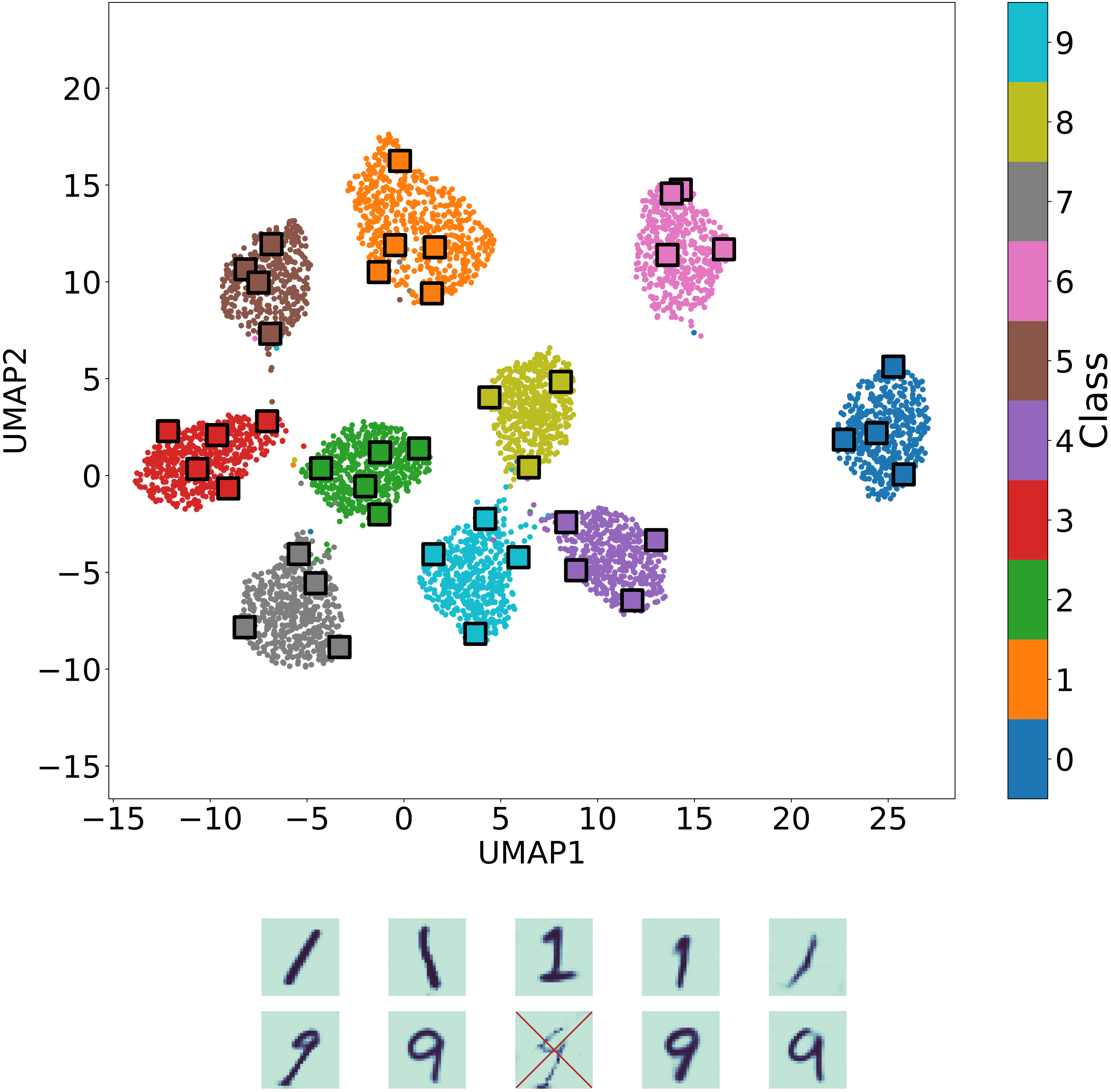}                   
        \caption{PanVAE.}
    \end{subfigure}
\caption{ProtoVAE (a) and PanVAE (b) visualizations of the latent space and decoded prototypes learned on MNIST after 30 epochs of training. Top: UMAP representations of the latent space with learned prototypes overlaid as squares. Bottom: Decoded prototypes of class '1' and '9'. One of the prototypes from PanVAE does not have the maximal similarity for any training image, indicated by a red cross. PanVAE has captured variations in the digit '1' pertaining to right-handedness (first '1' from the left), left-handedness (second '1' from the left) and a traditional writing style (third '1' from the left).}\label{fig:UMAP_MNIST30}
\end{figure*}

\begin{figure*}[ht!]
    \centering
    \begin{subfigure}[t]{1\columnwidth}
        \includegraphics[width=1\columnwidth]{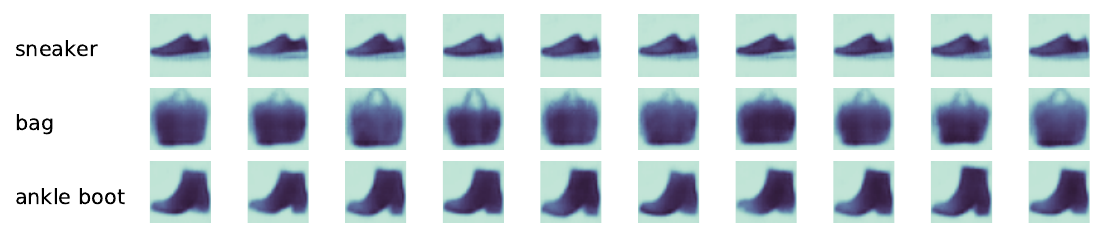}
        \caption{ProtoVAE $\mathcal{L}_\mathrm{orth}$ 1.}
        \label{fig:DivControl_orth1}
    \end{subfigure}
    ~ 
    \begin{subfigure}[t]{1\columnwidth}
        \includegraphics[width=1\columnwidth]{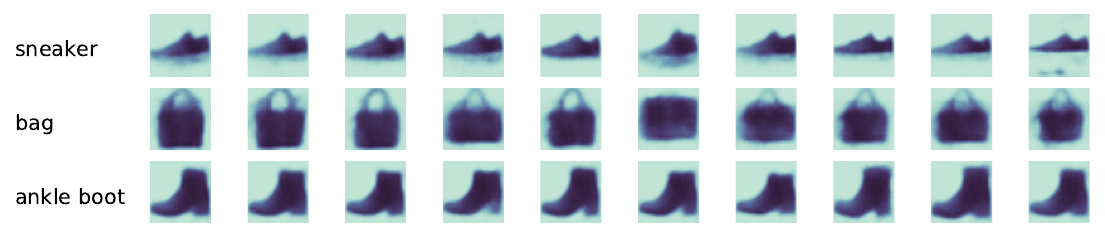}
        \caption{PanVAE $\mathcal{L}_\mathrm{vol}$ 1.}
        \label{fig:DivControl_vol1}
    \end{subfigure}
    \begin{subfigure}[t]{1\columnwidth}
        \includegraphics[width=1\columnwidth]{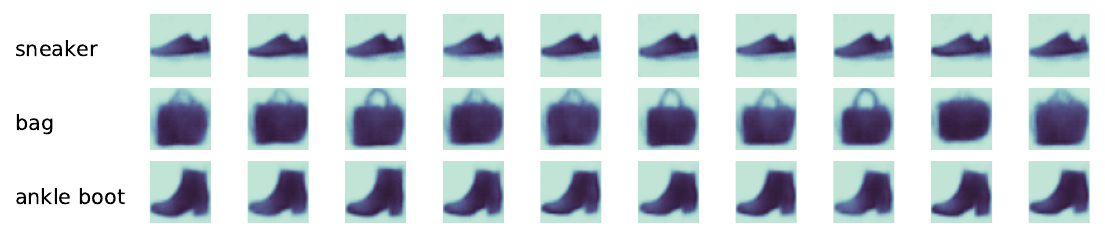}
        \caption{ProtoVAE $\mathcal{L}_\mathrm{orth}$ 100.}
        \label{fig:DivControl_orth100}
    \end{subfigure}
    \begin{subfigure}[t]{1\columnwidth}
        \includegraphics[width=1\columnwidth]{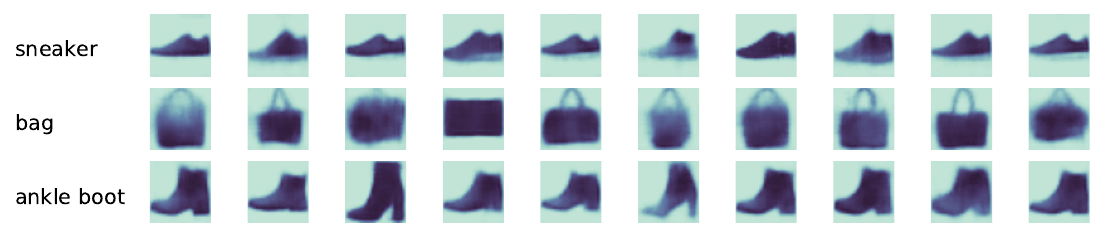}
        \caption{PanVAE $\mathcal{L}_\mathrm{vol}$ 100.}
        \label{fig:DivControl_vol100}
    \end{subfigure}    
    \caption{Diversity control enabled by ProtoVAE and PanVAE. The figure shows the change in decoded prototype appearance as the respective diversity inducing losses are increased. The prototypes are shown for the FMNIST data of classes "sneaker", "bag" and "ankle boot" after 10 epochs of training.  Figs. \ref{fig:DivControl_orth1} and \ref{fig:DivControl_orth100} show the difference between ProtoVAE prototypes with scale factor of 1 and 100 on the diversity loss $\mathcal{L}_\mathrm{orth}$. Figs. \ref{fig:DivControl_vol1} and \ref{fig:DivControl_vol100} show the difference between PanVAE pantypes with scale factor of 1 and 100 on the diversity loss $\mathcal{L}_\mathrm{vol}$.}\label{fig:DivControl}
\end{figure*}

\section{Results} \label{sec-res}
We perform experiments across various real-world datasets to monitor the transparency, diversity and trustworthiness of PanVAE. These datasets are FashionMNIST (FMNIST) \cite{xiao2017fashionFMNIST}, MNIST \cite{lecun1998gradientMNIST}, QuickDraw (QDraw) \cite{ha2017neuralQuickdraw} and CelebA \cite{liu2015CelebA}. We demonstrate the trustworthiness of PanVAE by evaluating the predictive performance of the overall model and asses the diversity and transparency using qualitative assessments from visualizations of the input space, as well as quantitative measures of prototype quality and coverage. We compare PanVAE to the performance of ProtoVAE and ProtoPNet.

\subsection{Predictive Performance}
The results for the predictive performance are shown in Tab. \ref{tab:Accuracy}, which demonstrates that PanVAE, like ProtoVAE, achieves higher predictive performance than ProtoPNET on the four datasets. There is no significant predictive performance gap between PanVAE and ProtoVAE on the datasets. This underlines the trustworthiness of PanVAE.

\begin{table}[t]
\begin{center}
\begin{small}
\begin{sc}
\begin{tabular}{lccr}
\toprule
Dataset & ProtoPNet & ProtoVAE  & PanVAE   \\
\midrule
MNIST             & 98.8 $\pm$ 0.1  &  \textbf{99.3 $\pm$ 0.1}  &   \textbf{99.4 $\pm$ 0.1}    \\
FMNIST            & 89.9 $\pm$ 0.5  & 91.6 $\pm$ 0.1            &    \textbf{92.2 $\pm$ 0.1}   \\
QDraw         & 58.7 $\pm$ 0.0  &  \textbf{85.6 $\pm$ 0.1}  &   \textbf{85.5 $\pm$ 0.1}    \\
CelebA            & 98.2 $\pm$ 0.1  & \textbf{98.6 $\pm$ 0.0}   &   \textbf{98.6 $\pm$ 0.0}     \\

\bottomrule
\end{tabular}
\end{sc}
\end{small}
\end{center}
\caption{Predictive performance (accuracy) of PanVAE ProtoVAE and ProtoPNet on MNIST, FMNIST, QuickDraw and CelebA. The values are the mean and standard deviation of three runs.}
\label{tab:Accuracy}
\end{table}

\subsection{Prototype Representation Quality}
Firstly, we asses prototype representation quality using visual inspection of the learned prototypes and the associated latent space. This can be seen for the MNIST dataset on Fig. \ref{fig:UMAP_MNIST30}, where the prototypes for ProtoVAE and PanVAE are shown. The diversity of PanVAE is higher than ProtoVAE. The prototypes from ProtoVAE are mostly orthogonal in latent space, but only occupy a small region of the space. Contrarily, the volume loss in PanVAE has pushed the pantypes away from each other allowing them to occupy and represent diverse regions of the dataspace. This is reflected in the decoded prototypes, which show high diversity by representing various archetypical ways of drawing digits. For instance, the pantypes capture variations between left-handed and right-handed digits of "1" as well as the archetypical "1" with a horizontal base. Moreover, PanVAE has found that the digits of "9" express less diversity and has thus pushed one of the pantypes OOD (indicated by a red cross in the figure). This form of prototype pruning by PanVAE allows the model to asses and represent the individual diversity expressed by each class. 

Fig. \ref{fig:DivControl} demonstrates the diversity control enabled by PanVAE by illustrating learned prototypes on the FMNIST datasets with different diversity loss scalings. The objective of the orthonormalization loss in ProtoVAE is to enforce intra-class diversity, and hence that the prototypes capture different concepts. While the loss ensures this, it only does so after sufficient training time. Fig. \ref{fig:DivControl} shows that scaling the orthonormalization loss in ProtoVAE does not significantly alter the diversity of the representation. On the other hand, the volumetric loss in PanVAE allows direct control over the diversity of the representation. 

\begin{figure*}[ht!]
    \centering
    \begin{subfigure}[b]{0.66\columnwidth}
        \includegraphics[width=1\columnwidth]{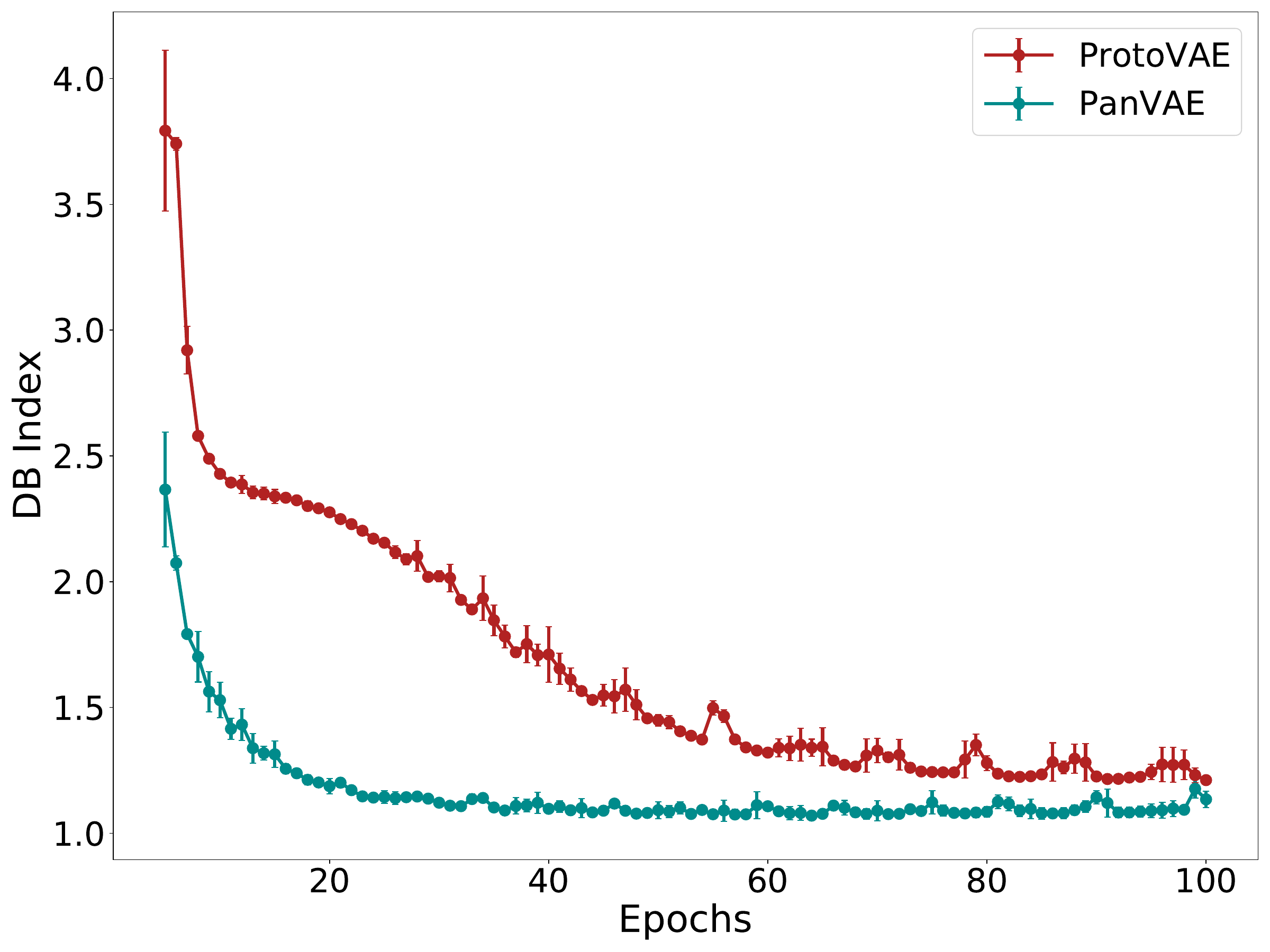}
        \caption{MNIST.}
        \label{fig:DB_MNIST}
    \end{subfigure}
    ~ 
    \begin{subfigure}[b]{0.66\columnwidth}
        \includegraphics[width=1\columnwidth]{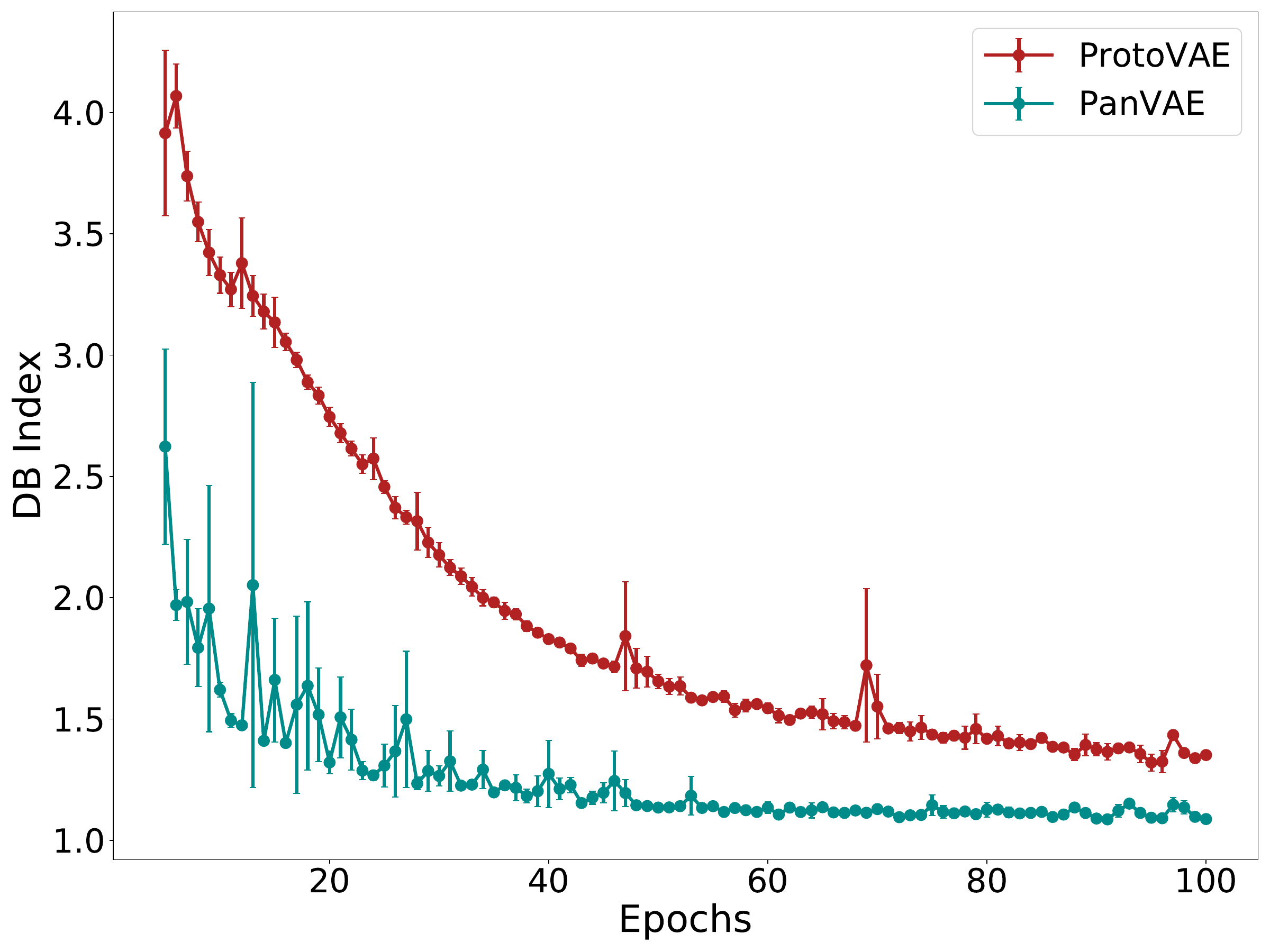}
        \caption{FMNIST.}
        \label{fig:DB_FMNIST}
    \end{subfigure}
    \begin{subfigure}[b]{0.66\columnwidth}
        \includegraphics[width=1\columnwidth]{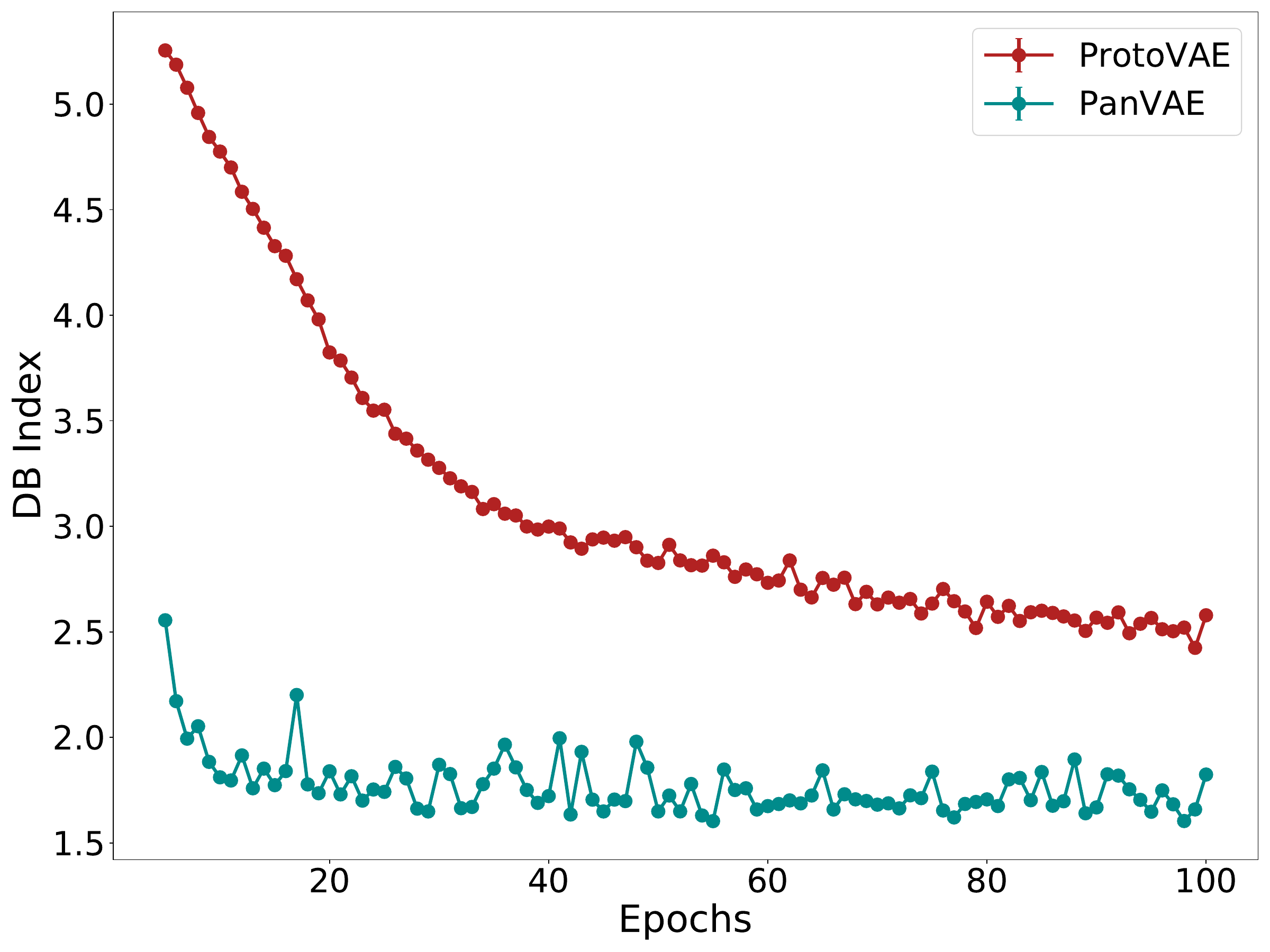}
        \caption{QuickDraw.}
        \label{fig:DB_QUICKDRAW}
    \end{subfigure}
    \caption{Evolution of prototype DB scores for PanVAE and ProtoVAE on MNIST, FMNIST and QuickDraw. Data points indicate mean values and associated standard deviations over three runs.}\label{fig:DB_Evolution}
\end{figure*}

Previous work in the literature on prototype based self-explainable classifiers often only qualitative asses the prototype diversity axis \citep{gautam2022protovae} (i.e. visual inspection of the diversity prerequisite of non-overlapping prototypes). We propose that self-explainable classifiers should not only be assessed with quantitative measures on the trustworthiness axis, but should also be evaluated by quantitative measures on the diversity axis. This includes thorough evaluations of how well the prototypes represent the dataspace. In order to do this we make use of measures of prototype quality and representativity by firstly measuring the prototype quality using the Davies-Bouldin (DB) index \citep{davies1979cluster} and secondly evaluating the diversity of the class representatives by assessing their data coverage.

\subsubsection{Davies-Bouldin Index}
The DB index is a measure of cluster quality defined by the average similarity between cluster $C_i$ for $i = 1,...,k$ and its most similar cluster $C_j$. The similarity measure $R_{ij}$ quantifies a balance between inter- and intra-cluster distances. We adopt this measure and consider the prototypes in a SEM as cluster representatives and assign observations to their closest prototype in latent space according to maximal similarity scores. The intra-cluster size $s_i$ is then measured as the average distance between prototype $i$ and each data point belonging to the prototype, while the the inter-cluster distance $d_{ij}$ is measured by the distance between prototypes $i$ and $j$. From this the cluster similarity measure $R_{ij}$ can be constructed such that it is non-negative and symmetric by:

\begin{equation}
    R_{ij} = \frac{s_i + s_j}{d_{ij}}.
\end{equation}

\noindent With these definitions in place the DB index may be defined by:

\begin{equation}
    DB = \frac{1}{k} \sum^{k}_{i=1}{\underset{i = j}{\mathrm{max}} \ R_{ij}},
\end{equation}
where a lower DB scores equates to a better representation of the underlying data. The DB scores for the different models can be seen in Tab. \ref{tab:DBscore}. PanVAE achieves the best DB scores in all cases, demonstrating the ability of the pantypes to represent the underlying dataspaces.

\begin{table}[t]
\begin{center}
\begin{small}
\begin{sc}
\begin{tabular}{lccr}
\toprule
Dataset &  ProtoPNet & ProtoVAE & PanVAE   \\
\midrule
MNIST         &   2.20 $\pm$ 0.18    & 1.21 $\pm$ 0.00   &   \textbf{1.13 $\pm$ 0.03}    \\
FMNIST        &   3.43  $\pm$ 1.15   & 1.35 $\pm$ 0.01   &  \textbf{1.09 $\pm$ 0.01}      \\
QDraw     &  2.52  $\pm$ 0.62    & 2.57 $\pm$ 0.01   &  \textbf{1.82 $\pm$ 0.01}     \\
CelebA        &  27.09  $\pm$ 27.23  & 1.58  $\pm$ 0.15  &  \textbf{1.37  $\pm$ 0.01}      \\
\bottomrule
\end{tabular}
\end{sc}
\end{small}
\end{center}
\caption{Davies-Bouldin scores of prototypes from the different models on the datasets used for our experiments. The values are the mean and standard deviation over three runs.}
\label{tab:DBscore}
\end{table}

In addition to achieving higher final DB scores, PanVAE also does so using less training time. This is illustrated in Fig. \ref{fig:DB_Evolution}, where the DB score evolution is shown for ProtoVAE and PanVAE over 100 epochs of training. PanVAE converges on a lower DB score much quicker than ProtoVAE. 

\begin{figure*}[ht!]
    \centering
    \begin{subfigure}[t]{1\columnwidth}
        \includegraphics[width=1\columnwidth]{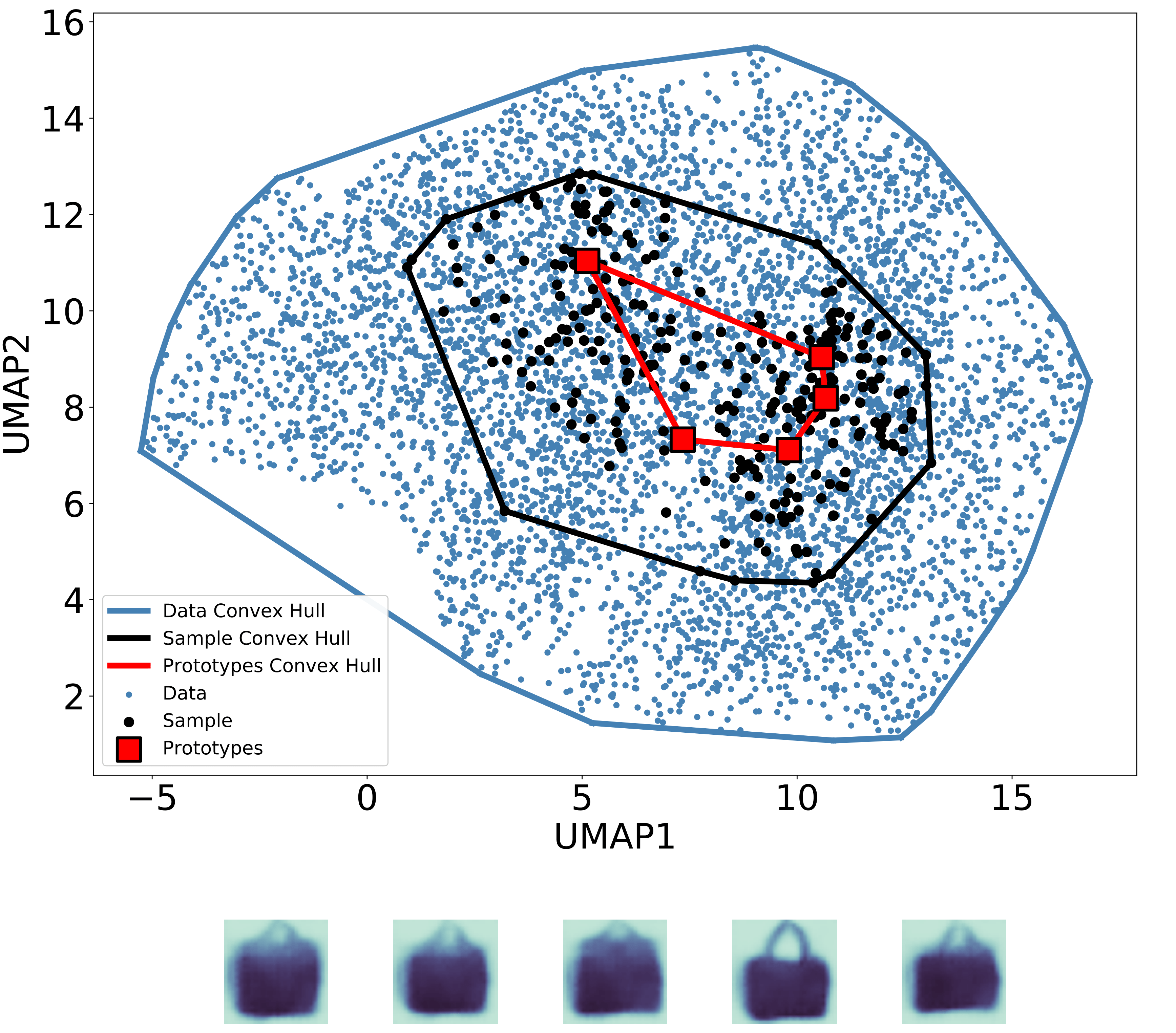}     
        \caption{Bag prototypes from ProtoVAE.}
    \end{subfigure}
    \begin{subfigure}[t]{1\columnwidth}    
        \includegraphics[width=1\columnwidth]{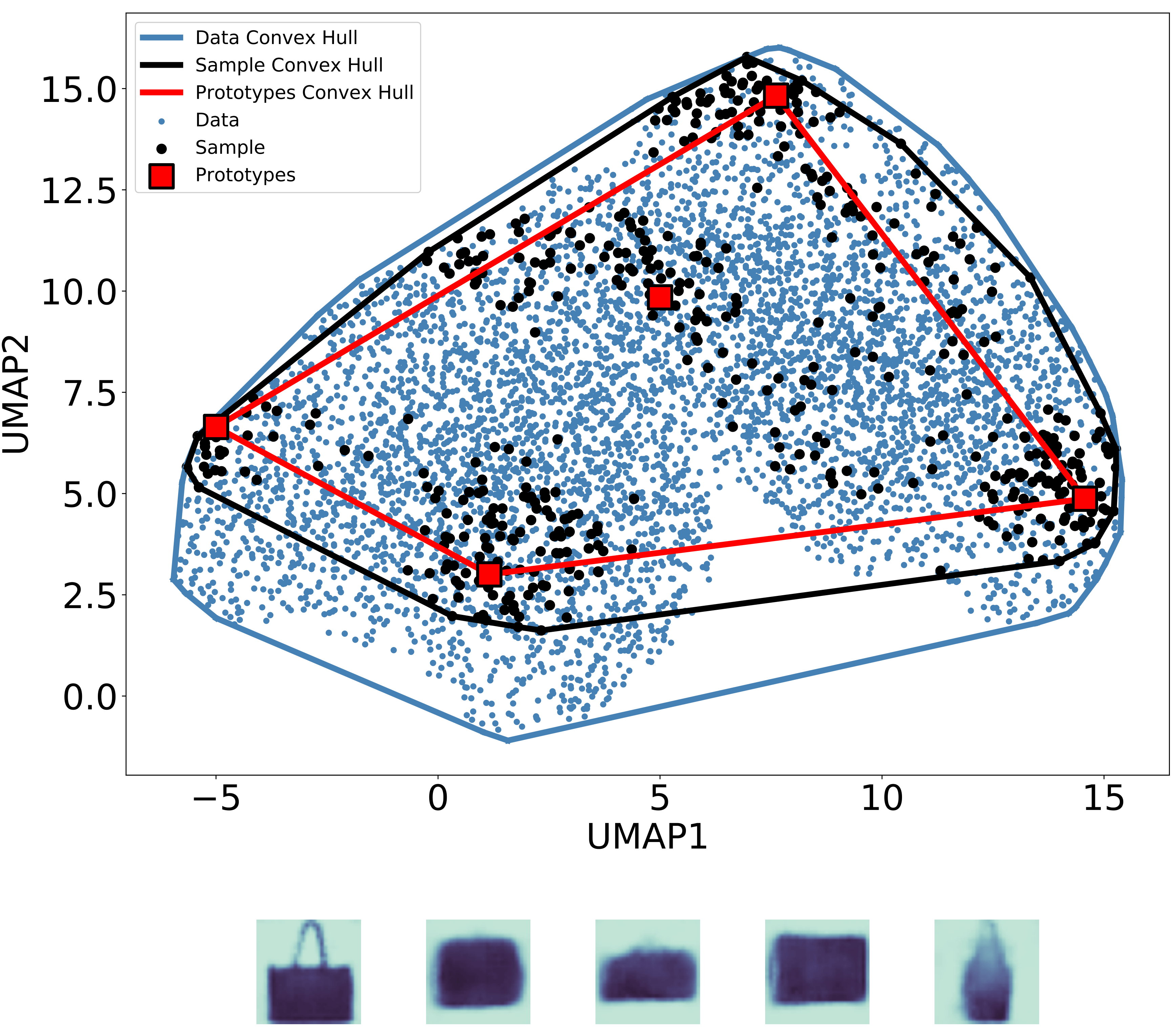}     
        \caption{Bag pantypes from PanVAE.}
    \end{subfigure}
    \caption{Prototype coverage in UMAP space from 20 epochs of training on FMNIST with 5 prototypes for the "bag" class for ProtoVAE (a) and PanVAE (b). Top: UMAP representations of the latent space with learned prototypes overlaid as red squares. The prototype convex hull in UMAP space is shown as a red outline around the prototypes and the full class dataspace convex hull is shown as a blue outline around the data. A sample of the 100 closest observations to each prototype is shown as black datapoints. The convex hull of the sampled observations is shown as a black outline. The PanVAE sample convex hull covers 77\% of the volume of the full class convex hull, whereas the ProtoVAE sample convex hull covers 33\%. Bottom: Decoded prototypes.}
    \label{fig:PrototypeCoverageFmnistBag}
\end{figure*}

\begin{figure*}[ht!]
    \centering
    \begin{subfigure}[b]{.8\linewidth}
        \includegraphics[width=1\columnwidth]{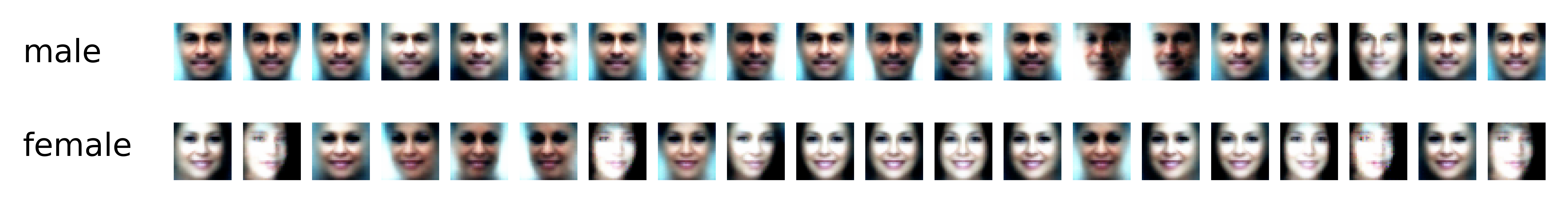}      
        \caption{ProtoVAE.}
    \end{subfigure}\\
    \begin{subfigure}[b]{.8\linewidth}    
        \includegraphics[width=1\columnwidth]{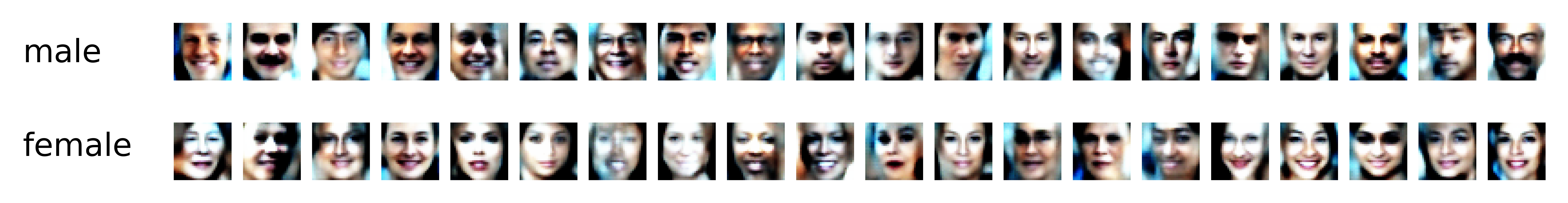}        
        \caption{PanVAE.}
    \end{subfigure}
    \caption{Face prototypes learned on the UTK Face dataset. The learned prototypes are shown for ProtoVAE in (a) and for PanVAE in (b). PanVAE has captured variations in race as well as other unseen features such as facial hair in males. The ProtoVAE males all have somewhat neutral expressions with shut mouths while most of the females have slight smiles. The PanVAE males and females all exhibit large variations in expression from full smiles with visible teeth to neutral expressions without visible teeth.}\label{fig:UTKFaces}
\end{figure*}

\subsubsection{Data Coverage}
The DB index provides a measure of prototype quality in terms of prototype representation quality, but does not sufficiently asses how well the prototypes cover the diversity in the dataspace. Sufficient coverage of various aspects in the dataspace has been found critical in obtaining unbiased ML algorithms \cite{jin2020mithracoverage}.

In order to asses prototype data coverage, we compare the volume spanned by observations represented by the prototypes to the volume of the full data distribution. Ideally, the prototypes are diverse enough, that they sufficiently cover a large volume of data they seek to represent. The coverage may be assessed through the volume of the convex hull of the data. We evaluate our pantypes on this premise by sampling the 100 nearest observations to each pantype. The proximity is measured in the full latent space in terms of the similarity score (Eq. \ref{EQ:sim_function}). We then compute the volume spanned by the represented observations from their convex hull, and compare this to the volume of the original data. We illustrate the results of this procedure in Fig. \ref{fig:PrototypeCoverageFmnistBag} using a 2D UMAP projection of the 256 dimensional latent space for the "Bag" class in FMNIST. The increased diversity of the pantypes allow them to occupy and represent a larger region of the dataspace.

\subsubsection{Demographic Diversity}
Sufficient representation of demographic groups has been found critical in ensuring ML fairness \cite{jin2020mithracoverage}. Image data used to train facial recognition algorithms have historically been biased towards White individuals, which are overrepresented in the training data, resulting in biased inference \cite{buolamwini2018genderdisparity}. The largest disparity is found between white skinned and dark skinned individuals. 

\begin{table}[h]
\begin{center}
\begin{small}
\begin{sc}
\begin{tabular}{lccr}
\toprule
Metric & ProtoVAE & PanVAE   \\
\midrule
Acc All                       &   \textbf{95.08 $\pm$ 0.11}  &  \textbf{95.42 $\pm$ 0.37}  \\
Acc White Male              &   \textbf{ 96.35 $\pm$ 0.31} & 95.21 $\pm$ 0.33     \\
Acc Black Female            &  91.67 $\pm$ 0.53 &  \textbf{94.90 $\pm$ 0.39}   \\
Acc Gap                    &   4.69 $\pm$ 0.24 &  \textbf{0.32 $\pm$ 0.15}     \\
Diversity           &   1.26 $\pm$ 0.06  & \textbf{1.43 $\pm$ 0.07}     \\
\bottomrule
\end{tabular}
\end{sc}
\end{small}
\end{center}
\caption{UTK results. The values are the mean and standard deviation of three runs. The overall accuracy is reported along with the individual accuracy and accuracy gap between White males and Black females. A positive gap value indicates that the mean accuracy is higher on White males compared to Black females. Diversity is the information entropy (demographic diversity) of the distribution of races represented by the prototypes. The represented races are determined by the nearest test image to each prototype.}
\label{tab:UTKResults}
\end{table}

Demographic diversity may be quantified using a measure of combinatorial diversity, also known as diversity index \cite{simpson1949measurement}. The combinatorial diversity is defined as the information entropy of the distribution \cite{celis2016fairdiverse}:
\begin{equation}
    H = -\sum\limits_{i=1}^k p_i \ \mathrm{log} \ p_i,
    \label{Eq:Entropy}
\end{equation}
where the combinatorial diversity measure $H$ is the entropy, $p_i$ is the probability of event $i$ and $\sum$ is the sum over the possible outcomes $k$. This measure quantifies the information entropy of the demographic distribution over $k$ demographic groups. A high entropy equates to a more diverse (fair) representation, which is not particularly biased towards any demographic group. 

We evaluate how the volumetric loss may aid in mitigating demographic data bias and enhance group level diversity. To do this we train PanVAE on the UTK Face dataset \cite{zhang2017ageUTKFACE}, which contain images of about 20,000 individuals with associated sex and race labels. The decoded facial prototypes from training on the UTK Face dataset can be seen in Fig. \ref{fig:UTKFaces}. To evaluate the demographic diversity, we asses the race of the nearest test image to each prototype and use this to compute the combinatorial diversity of the race distribution. The overall accuracy and diversity results are reported in Tab. \ref{tab:UTKResults}. We also report the accuracy gap between White males and Black females. This accuracy gap has been identified as a ubiquitous problem in facial recognition algorithms. White males account for 23 percent of the individuals in the UTK Face data, while Black females account for 9 percent. PanVAE achieves a lower accuracy gap between these demographics due to a better accuracy on Black females. However, this comes at the expense of a lower accuracy on the majority sub-population of White males as compared to ProtoVAE.

\section{Discussion}
The volumetric loss in PanVAE promotes the generation of diverse prototypes, which capture the underlying dataspace and represent distinct archetypical patterns in the data. This leads to increased representation quality and data coverage and can mitigate data bias. However, pantypes are most useful when the diversity expressed by the input data aligns with the diversity a study aims to enforce. This is closely related to the concepts of geometric and combinatorial diversity \cite{celis2016fairdiverse}, where geometric diversity expresses the volume spanned by a number of high-dimensional feature vectors and combinatorial diversity is related to information entropy of discrete variables. This means that geometric diversity is useful for ensuring what humans perceive as high \textit{visual} diversity, while combinatorial diversity is useful for ensuring high \textit{demographic} diversity (or fairness) of human understandable sensitive variables that take on a small number of discrete values (such as race). The volumetric loss in PanVAE exclusively ensures a large geometric diversity of the learned pantypes and as such only enforces visually diversity. This may not necessarily align with the diversity in unseen protected attributes such as race in facial image data. This misalignment can occur if features like background color and pose in the facial images exhibit larger visual variation than features related to demographic diversity such as skin tone. To enforce high demographic diversity, the images would either have to be pose aligned and background removed (or at at least background noise reduced) or the sensitive features would have to be incorporated directly into the model, if possible. We have trained PanVAE on the cropped and aligned version of the UTK Face dataset to demonstrate that geometric and combinatorial diversity can be obtained simultaneously in noise reduced data with the volumetric loss. More balanced demographic representation can lead to better predictive performance for minority sub-populations in the data and consequently less disparate predictive performance between sub-populations. However, this usually comes at the expense of a reduction in performance for the majority group. Thus, the choice of representation should be carefully considered in coherence with the aim and target population of the trained model.

\section{Conclusion}
We have introduced pantypes, a new family of prototypical objects used in a SEM to capture the full diversity of the dataspace. Pantypes emerge by virtue of a volumetric loss and are easily integrated into existing prototypical self-explainable classifier frameworks. The volumetric loss causes the pantypes to diverge early in the training process and to capture various archetypical patterns through a sparse set of objects leading to increased interpretability and representation quality without sacrificing accuracy.

\bibliography{refs}
\section{Acknowledgments}
We would like to acknowledge the authors of the well-tested ProtoVAE. We have used the public code for this model as the foundation of PanVAE.

\end{document}